\documentclass[conference,final]{IEEEtran}
\IEEEoverridecommandlockouts

\usepackage{graphicx}

\usepackage[maxbibnames=5,style=ieee]{biblatex}
\usepackage{url}

\usepackage{amsmath,amssymb,amsfonts}
\usepackage{bm}

\usepackage{textcomp}
\usepackage{fancyhdr}
\usepackage{transparent}

\usepackage{pifont}
\usepackage{tabularray}
\UseTblrLibrary{booktabs}
\usepackage{threeparttable}

\usepackage{wrapfig}

\usepackage{ifthen}
\newboolean{isshortieee}
\setboolean{isshortieee}{true}
\newboolean{isfancyieee}
\setboolean{isfancyieee}{false}

\DeclareMathOperator*{\argmax}{arg\,max}

\AtBeginBibliography{\footnotesize}

\def\BibTeX{{\rm B\kern-.05em{\sc i\kern-.025em b}\kern-.08em
    T\kern-.1667em\lower.7ex\hbox{E}\kern-.125emX}}

\fancypagestyle{copyright}{
    \fancyhf{}
    \fancyhead[C]{
        \small
        \textit{IEEE International Conference on Multimedia Information Processing and Retrieval (MIPR)}, San Jose, CA, 2024. \\
        \vspace{5pt} \scriptsize
        © 2024 IEEE. Personal use of this material is permitted. Permission from IEEE must be obtained for all other uses, in any current or future media, including reprinting/republishing this material for advertising or promotional purposes, creating new collective works, for resale or redistribution to servers or lists, or reuse of any copyrighted component of this work in other works.
    }
    
}

\usepackage{anyfontsize}

\usepackage{algorithm}
\usepackage[noend]{algpseudocode}

\begin{document}

\title{Retrieval Augmented Structured Generation:\\\fontsize{22}{24}\selectfont Business Document Information Extraction As Tool Use}



\author{\IEEEauthorblockN{Franz Louis Cesista, Rui Aguiar, Jason Kim, Paolo Acilo}
\IEEEauthorblockA{\textit{Expedock Software, Inc}, San Francisco, United States\\
\tt\footnotesize franzlouiscesista@gmail.com, raguiar1000@gmail.com, jasonminsookim@gmail.com, paolo@expedock.com}
\vspace{-0.5cm}
}

\maketitle
\thispagestyle{copyright}

\begin{abstract}
Business Document Information Extraction (BDIE) is the problem of transforming a blob of unstructured information (raw text, scanned documents, etc.) into a structured format that downstream systems can parse and use. It has two main tasks: Key-Information Extraction (KIE) and Line Items Recognition (LIR). In this paper, we argue that BDIE is best modeled as a \textit{Tool Use} problem, where the tools are these downstream systems. We then present Retrieval Augmented Structured Generation (RASG), a novel general framework for BDIE that achieves state of the art (SOTA) results on both KIE and LIR tasks on BDIE benchmarks.

The contributions of this paper are threefold: (1) We show, with ablation benchmarks, that Large Language Models (LLMs) with RASG are already competitive with or surpasses current SOTA Large Multimodal Models (LMMs) without RASG on BDIE benchmarks. (2) We propose a new metric class for Line Items Recognition, General Line Items Recognition Metric (GLIRM), that is more aligned with practical BDIE use cases compared to existing metrics, such as ANLS*, DocILE, and GriTS. (3) We provide a heuristic algorithm for backcalculating bounding boxes of predicted line items and tables without the need for vision encoders. Finally, we claim that, while LMMs might sometimes offer marginal performance benefits, LLMs + RASG is oftentimes superior given real-world applications and constraints of BDIE.
\end{abstract}




\section{Introduction}

In practice, Business Document Information Extraction (BDIE) is done because one wants to connect human organizations such as businesses and governments with indefinite interfaces (such as printed/handwritten documents) to downstream systems with definite interfaces (such as Application Program Interfaces). For a machine learning system to be good at BDIE, it has to be good at interfacing with these downstream programs and at least minimally aware of what these programs do with the data. In the succeeding sections, we discuss how we can achieve these goals with our new methodology, Retrieval Augmented Structured Generation (RASG), and our new proposed metric class, General Line Items Recognition Metric (GLIRM).



BDIE has two main sub-problems, Key Information Extraction (KIE) and Line Items Recognition (LIR) \cite{skalický2022business}.
The goal of KIE is to extract and format information from the document into key-value pairs. And the goal of LIR is to extract information into a list of line items where each line item corresponds to a row in a table formatted into column key-value pairs.
And unlike Table Structure Recognition, the order of the columns do not matter so long as the columns are mapped to the proper, predefined column keys \cite{smock2023grits}.

To summarize and motivate RASG: the goal of BDIE is to transform a blob of information into a structured format to pass to downstream tools (i.e. APIs). We can teach machine learning models to use tools through supervised finetuning. We can then force them to output in the format we expect using structured generation. We can also teach sufficiently-powerful, pretrained models to use new tools on out-of-distribution datasets by taking advantage of in-context learning. Finally, to be able to use commercial LLMs "out-of-the-box", we can structure the text prompt to \textit{look} like the original document through prompt engineering.



\section{Retrieval Augmented Structured Generation}
\ifthenelse{\boolean{isfancyieee}}{}{
\begin{figure*}[tb]
\centerline{\includegraphics[scale=0.20]{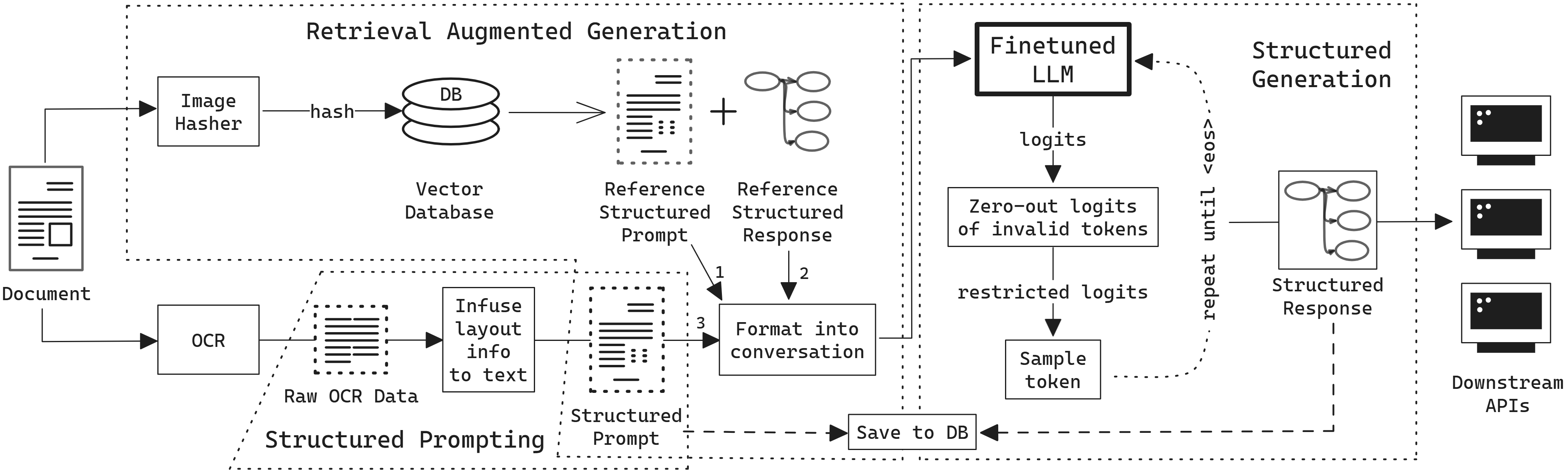}}
\caption{\textbf{Retrieval Augmented Structured Generation (RASG).} We model Business Document Information Extraction as a \textit{Tool Use} problem with downstream APIs as the tools. We then combine Retrieval Augmented Generation, Supervised Finetuning, \& Structured Generation, techniques that improve tool use capabilities of ML models, with Structured Prompting to beat strong multimodal models using only LLMs. This allows to use the largest open-source and commercial models to reach SOTA results at minimal costs.}
\label{fig1}
\vspace{-10pt}
\end{figure*}
}
Retrieval Augmented Structured Generation (RASG) is composed of four components: \textbf{(1) Retrieval Augmented Generation} which allows us to teach LLMs to use new tools using In-Context Learning \cite{lewis2021retrievalaugmented}; \textbf{(2) Supervised Finetuning} which enhances the correctness of the extracted outputs; \textbf{(3) Structured Generation} which ensures that the outputs are parse-able by downstream programs \cite{willard2023outlines}; and \textbf{(4) Structured Prompting} which infuses layout information into the prompt \cite{wang2023latin}.

All four components are necessary to beat strong multimodal baselines on BDIE using an open-source 7B LLM, Hermes 2 Pro - Mistral 7B \cite{anon2024hermespro}. But only a subset are necessary when using GPT-3.5 \cite{openai-gpt35-docs}.

\subsection{Notes for Finetuning for Structured Generation}

The language model used must output both the right \textit{content} and the right \textit{structure} of said content. Finetuning significantly helps with the former, less so with the latter. To ensure that the output is parseable by downstream systems, we need to zero the probabilities of invalid tokens. This is where structured generation works. Based on our experiments, we found that naively combining finetuning and structured generation leads to poor results. There are two main issues:

1. \textbf{Schema vs. model mismatch}: Regex-based algorithms for structured generation, such as Outlines' \texttt{outlines.generate.json} module, implicitly impose a strict key ordering \cite{willard2023outlines} \cite{tang2024loraxoutlines}. E.g., suppose that we have a schema where the key \texttt{"amount"} comes before \texttt{"currency"}. Then, Outlines will mask the logits for \texttt{"currency"} until \texttt{"amount"} is generated. However, if the model was finetuned to generate \texttt{"currency"} before \texttt{"amount"}, the prediction accuracy collapses. To remedy this, one must either ensure that the finetuning dataset strictly follows the specified schema, or use Context-Free Grammar-based algorithms for structured generation, such as Outlines' \texttt{outlines.generate.cfg} module, which does not impose a strict key ordering.

2. \textbf{Token explosion with optional keys}. A common issue we have observed is to require the keys to be generated even when the predicted value is \texttt{null}. E.g., when one builds a Pydantic object with \texttt{Optional} fields then naively pipes the object's json schema to Outlines. This leads to a lot of unnecessary tokens being generated, slowing down inference. Another bad practice when using Outlines is to make \textit{all} of the keys optional. This is because Outlines uses a different algorithm to generate the FSM for this case. A workaround for this is to add a required dummy key of type \texttt{null} to the schema and remove it in postprocessing.

\subsection{Bounding Box Backcalculation Heuristic}
For the KIE task, we have found that a simple, greedy algorithm (Algo \ref{bboxheuristic}) suffices for backcalculating bounding boxes. To use the entire page, simply set the $y$ lowerbound and upperbound to $0$ and the page height in pixels, respectively. For the LIR task, a good heuristic is to (1) divide the page vertically into chunks, one for each line item; and (2) re-use Algorithm 1 above to back-calculate the bounding boxes for each line item, but only for the words in the chunk assigned to the line item. The challenge is \textit{how} to divide the page.
{
\small
\begin{algorithm}[b]
\footnotesize
\caption{Bounding Box Backcalculation Heuristic}\label{alg:bboxheuristic}
\begin{algorithmic}[1]
\Require $y$ lower- \& upperbound, predicted key-value map, and OCR data
\Ensure Matching score, key to bounding box mapping
\State $score \gets 0; key\_bbox\_map \gets \{\}$
\ForAll{$(key, value)$ pair in the key-value map}
    \State $m\_words \gets$ The longest contiguous list of words matching $value$ whose bounding boxes lie within the $y$ lowerbound and upperbound
    \State $m\_bboxes \gets \text{Bounding boxes of } m\_words$
    \State $key\_bbox\_map[key] \gets union(m\_bboxes)$\;
    \State $score$ $\gets$ $score + similarity(concat(m\_words)), value)$
\EndFor
\State \textbf{return} $(score, key\_bbox\_map)$
\end{algorithmic}
\label{bboxheuristic}
\end{algorithm}
}

The naive dynamic programming approach with 2D states \texttt{(line item index, page $y$)} has complexity $O(MN^2 \cdot O(\text{Algo \ref{bboxheuristic}}))$ where $M$ is the number of line items and $N$ is the page height. We can optimize this by down-scaling the page. In production we use $N = 128$. A further optimization takes advantage of the monotonicity of Algorithm 1: the matching score is non-increasing as we increase the $y$ lower bound and decrease the $y$ upper bound.
Therefore, we can use the Divide-and-Conquer optimization for dynamic programming problems to speed this up to $O(MN\log{N} \cdot O(\text{Algo \ref{bboxheuristic}}))$. Finally, we employ binary search to find the largest $y$ lowerbound for the first and the smallest $y$ upperbound for the last line item to tighten the bounds.

\section{General Line Items Recognition Metric}

The goal of LIR is to extract information into an ordered list of line items where each line item corresponds to a row in a table and is formatted into column key-value pairs. In this section, we will derive a new metric class for LIR. 

A metric for LIR should have the following attributes:

{ \small
\begin{enumerate}
 \item  \textbf{Subtask Isolation}: Performance on the subtasks must be measured separately.
 \item  \textbf{Cell Isolation}: A True Positive corresponds to exactly one predicted cell and one ground truth cell.
 \item  \textbf{Cell Completeness}: Hallucinated cells are counted as False Positives. Missing cells are counted as False Negatives.
 \item  \textbf{Cell Similarity-Measure Flexibility}: Within the same subtask, the metric must support multiple cell similarity measures.
 \item  \textbf{Cell Row-Position Invariance}: The credit given to a correctly predicted cell is the same regardless of absolute row position.
 \item  \textbf{Row-Order Preservation}: For any two predicted rows, their relative order and the relative order of their matching ground truth rows must be the same.
 \item  \textbf{Column-Permutation Invariance}: The metric must be invariant to column shuffling.

\end{enumerate}

}
Attribute \#1 is important because we do not often need both the cell content and location information for downstream tasks. Attributes \#2 and \#3 are so that we can have an F1-score-like metric that penalizes both over-extraction and under-extraction of cells. Attribute \#4 is so the metric can be extended to support multiple similarity measures more appropriate for downstream tasks.

Attribute \#5 is relevant in cases where there are extra or missing rows in the predictions. We only want to penalize such rows, not the rows with matching ground truth. Attribute \#6 is important because there are documents where the order of the rows matter. For example, documents where transactions are ordered by time.

And finally, Attribute \#7 is an \textit{ideal} attribute because the order of the columns do not often matter when applying business logic to the line items. Downstream programs often then save the results (and the extracted data) to column-invariant SQL-like databases. 

\subsection{Limitations of Current Line Items Recognition Metrics}

ANLS* and the DocILE metric use Maximum-Weight Bipartite Matching-based algorithms for row-matching \cite{peer2024anls} \cite{simsa2023docile}. Thus, they do not satisfy attribute \#6. Furthermore, the latter supports both cell content and cell location recognition but does not isolate the two--violating attribute \#1. This makes it impossible to use for cell content recognition only or cell location recognition only tasks. GriTS satisfies all of the attributes above except for attribute \#7 \cite{smock2023grits}.

For the rest of the section, we describe a new metric that satisfies all of the attributes above which we call General Line Items Recognition Metric (GLIRM). This metric can both be viewed as an extension of ANLS* and DocILE so they satisfy attributes \#1 and \#6, and a relaxation of GriTS, so it satisfies attribute \#7.

\subsection{Similarity Matching Score}

As per attributes and \#1 and \#4, we will use $f(c_p, c_t)$ to denote the similarity measure between a predicted cell, $c_p$, and a ground truth cell, $c_t$. $f$ can be any similarity measure appropriate for the downstream task. Exact match for product reference numbers, Intersection-over-Union for bounding boxes, etc. To make the metric F1-score-like, we have to constrain $f$ to be between $0$ and $1$: $0 \le f(c_p, c_t) \le 1$, for all $c_p, c_t$. We will use use $g_f(row_p, row_t)$ as the sum of the similarity scores of the corresponding cells in the predicted row, $row_p$, and the ground truth row, $row_t$.

\subsection{Row Matching}

Let's denote $R_p$ and $R_t$ as the sequence of rows in the predicted and ground truth line items, respectively. We then denote $R'_p$ and $R'_t$ as \textit{subsequences} of rows in  $R_p$ and $R_t$, respectively. The goal is to find equal-length subsequences $\tilde{R}_p$ and $\tilde{R}_t$ such that the sum of the similarity scores of the corresponding cells is maximized:
{\small
\vspace{-5pt}
\begin{equation}
    \tilde{R}_p,\tilde{R}_t = \argmax\nolimits_{R'_p | R_p, R'_t|R_t} \sum\nolimits_{i} g_f(R'_p[i], R'_t[i])
    \vspace{-5pt}
\end{equation}
}
Because we concern ourselves with \textit{subsequences} instead of subsets of rows, it is more appropriate to use a Levenshtein Distance-like algorithm to find $\tilde{R}_p,\tilde{R}_t$ rather than a Maximum-Weight Bipartite Matching-based algorithm as in ANLS* and DocILE. This penalizes swapped or shuffled rows in the predictions.

This approach is similar to the one used in the GriTS metric, but in one dimension instead of two, and does not enforce column-order preservation.

\subsection{General Line Items Recognition Metric}

To define the General Line Items Recognition Metric (GLIRM), we first define the GLIRM-Precision and GLIRM-Recall scores as follows:
{\small
\vspace{-5pt}
\begin{equation}
    \text{GLIRM-Prec}_f(R_p, R_t) = (1/|R_t|)\sum\nolimits_{i} g_f(\tilde{R}_p[i], \tilde{R}_t[i])
    \vspace{-5pt}
\end{equation}
\vspace{-5pt}
\begin{equation}
    \text{GLIRM-Rec}_f(R_p, R_t) = (1/|R_p|)\sum\nolimits_{i} g_f(\tilde{R}_p[i], \tilde{R}_t[i])
    \vspace{-5pt}
\end{equation}
}
The F1-score-like GLIRM then is,
{\small
\vspace{-5pt}
\begin{equation}
    \text{GLIRM-F1}_f(R_p, R_t) = \dfrac{2 \sum_{i} g_f(\tilde{R}_p[i], \tilde{R}_t[i])}{|R_p| + |R_t|}
    \vspace{-5pt}
\end{equation}
}
In practice or if humans are reviewing the output of the system, recall is often more important than the precision. This is because it takes more time to look for and box-in missing cells than to verify the correctness of the extracted cells. Thus, we can define GLIRM as:
{\small
\vspace{-5pt}
\begin{equation}
    {{\text{GLIRM-F1}_{\beta}}_f}(R_{p}, R_{t}) = \dfrac{(1 + \beta^2) \sum_{i} g_f(\tilde{R}_p[i], \tilde{R}_t[i])}{\beta^2 |R_p| + |R_t|}
    \vspace{-5pt}
\end{equation}
}
where $\beta$ is a hyperparameter that controls the importance of the recall over the precision. If $\beta = 1$, then the metric is the same as $\text{GLIRM-F1}_f$.

\section{Experiments}

\subsection{Dataset}
We used the DocILE dataset for benchmarking \cite{simsa2023docile}. This dataset is a large-scale research benchmark for machine learning evaluation of Key Information Extraction (KIE) and Line Item Recognition (LIR) from semi-structured business documents such as invoices.

\subsection{Methods}
We ablated three components of RASG: Retrieval Augmented Generation, Supervised Finetuning, and Structured Prompting. We did not include Structured Generation in the ablation benchmarks because it is a necessary component for BDIE. Without Structured Generation, we will not be able to guarantee that the output of the model is interpretable by downstream systems. We also ran experiments with two types of models: a commercially-available LLM, GPT-3.5, and an open-source LLM, Hermes 2 Pro - Mistral 7B. 

For Supervised Finetuning, we used OpenAI's Finetuning API for GPT-3.5 while we used 8Bit QLoRA on Axolotl to finetune Hermes 2 Pro - Mistral 7B on a single 80GB A100 GPU \cite{dettmers2023qlora} \cite{axolotl}. For the retrieval mechanism, we measured the "similarity" between pages using the manhattan distance of their wavelet hashes \cite{sing2017wavelethash}. For Structured Generation, we used OpenAI's Tool Use API for GPT-3.5 while we used Outlines for Hermes 2 Pro - Mistral 7B \cite{willard2023outlines}. Finally, we used LATIN-prompt for Structured Prompting \cite{wang2023latin}.

We finetuned the models for only one epoch and only used one-shot retrieval instead of many-shot retrieval primarily due to token window limits. Business documents are often dense, and a context window containing multiple documents is too large for current language models.

Overall, we ran a total of $2^4 = 16$ experiments; one for each combination of the components and base model. We then fitted a linear model to determine the contribution of each component to the overall performance of the model.

\subsection{Results}

Table I compares the performance of LLMs with RASG on the KIE and LIR tasks, respectively, against strong, multimodal LayoutLMv3 and Roberta + DETR baselines \cite{vimsa2023docileExtended} \cite{huang2022layoutlmv3} \cite{liu2019roberta} \cite{carion2020detr}. Table II shows the individual contribution of each component of RASG by base model.

The minimal resources needed to beat the baselines on the KIE task are either GPT-3.5 + 1-Shot Retrieval or Hermes 2 Pro + full RASG if one is required to run inference using open source components. For the LIR task, GPT-3.5 + 1-Shot Retrieval + Structured Prompting is sufficient to beat the baselines.

Finally, we measured the median table-level Information Coverage Score (ICS) for the bounding box backcalculation heuristic \cite{xiao2023informationCoverageScore}. The best baseline, Roberta + finetuned DETR, achieves 92.93\% ICS while GPT-3.5 + RASG and Hermes 2 Pro + RASG achieves 87.79\% and 85.02\% ICS, respectively.

{
\def\arraystretch{0.75}
\begin{table}[t]
\centering
\footnotesize
\caption{Model performance benchmarks on KIE \& LIR tasks on the DocILE dataset}
\begin{threeparttable}
\begin{booktabs}{lrr}
    \toprule
    \textbf{Model} & \textbf{KIE F1 Score} & \textbf{LIR GLIRM-F1}\\
    \midrule
    LayoutLMv3 & $63.95\%$ & $70.12\%$ \\
    LayoutLMv3 + synthetic data\tnote{1} & $65.28\%$ & $71.02\%$ \\
    \midrule[dotted]
    Roberta & $65.88\%$  & $70.44\%$ \\
    Roberta + synthetic data\tnote{1} & $\bm{65.97\%}$  & $71.83\%$ \\
    Roberta + DeTr Line Items\tnote{2} & $\cdots$ & $49.56\%$ \\
    Roberta + DeTr Table\tnote{3} & $\cdots$ & $\bm{72.73\%}$ \\
    \midrule
    Hermes 2 Pro                                       & 16.44\% & 7.06\% \\
    Hermes 2 Pro + RASG & $\bm{73.41\%}$ & $69.44\%$ \\
    \midrule[dotted]
    GPT-3.5             & $26.80\%$      & $23.03\%$ \\
    GPT-3.5 + RASG      & $\bm{75.40\%}$ & $\bm{79.81\%}$ \\
    \bottomrule
\end{booktabs}
\begin{tablenotes}
    \item[1]Additional pretraining on synthetic data provided by the DocILE dataset.
    \item[2]Detection Transformer (DeTr) finetuned on line items detection.
    \item[3]Detection Transformer (DeTr) finetuned on table detection.
\end{tablenotes}
\end{threeparttable}
\vspace{-10pt}
\end{table}
}

{
\def\arraystretch{0.75}
\begin{table}[t]
\footnotesize
\centering
\caption{Ablation Benchmarks of RASG components on KIE \& LIR tasks on the DocILE dataset}
\begin{tabular}{lrr}
    \toprule
    \textbf{Model} & \textbf{KIE F1 Score} & \textbf{LIR GLIRM-F1} \\
    \midrule
    \textbf{GPT-3.5}        &   $34.17\%$ &   $28.31\%$ \\
    + 1-Shot Retrieval      & + $22.08\%$ & + $20.67\%$ \\
    + Supervised Finetuning & + $22.31\%$ & + $17.73\%$ \\
    + Structured Prompting  & + $4.96\%$  & + $19.42\%$ \\
    \midrule
    \textbf{Hermes 2 Pro - Mistral 7B} & $13.55\%$ & $4.69\%$ \\
    + 1-Shot Retrieval      & + $36.87\%$ & + $40.55\%$ \\
    + Supervised Finetuning & + $17.71\%$ & + $13.53\%$ \\
    + Structured Prompting  & +  $0.63\%$ & + $10.30\%$ \\
    \bottomrule
\end{tabular}
\vspace{-10pt}
\end{table}
}



\section{Discussion and Conclusion}

Our model performance and ablation results demonstrate a few conclusions. Firstly, for KIE, prompt engineering only provides marginal gains compared to augmenting the model with a retrieval mechanism and/or finetuning it on the target dataset. For LIR, however, prompt engineering is as important as retrieval mechanisms and finetuning. Interestingly, properly tuned and augmented LLMs, can beat finetuned multimodal models such as LayoutLMv3 and Roberta + DeTr. Lastly, our bounding box backcalculation heuristic is only slightly worse than the best baseline at table detection despite not being optimized directly for the task.

For teams working in the Business Document Information space, our recommendation is to start with off-the-shelf LLMs that support structured generation, then implement a retrieval mechanism. If the performance is still poor, consider supervised finetuning. For LIR, we recommend starting with structured prompting first, then finetuning.




\printbibliography


\end{document}